\documentclass[letterpaper, 10 pt, conference]{ieeeconf}  

\IEEEoverridecommandlockouts                              

\usepackage{graphics} 
\usepackage{epsfig} 
\usepackage{mathptmx} 
\usepackage{times} 
\usepackage{amsmath} 
\usepackage{amssymb}  
\usepackage{float}
\usepackage[fleqn]{mathtools}
\usepackage{algorithm}
\usepackage[noend]{algpseudocode}
\makeatletter
\def\BState{\State\hskip-\ALG@thistlm}
\makeatother
\graphicspath{{figures/}}

\title{\LARGE \bf
A sub-optimal sampling based method for path planning
}

\author{Mahdi Morsali$^{1}$ and Fatemeh Mohseni$^{2}$
}

\begin{document}

\maketitle
\thispagestyle{empty}
\pagestyle{empty}

\begin{abstract}
In this paper a search algorithm is proposed to find a sub optimal path for a non-holonomic system. For this purpose the algorithm starts sampling the front part of the vehicle and moves towards the destination with a cost function. The bicycle model is used to define the non-holonomic system and a stability analysis with different integration methods is performed on the dynamics of the system. A proper integration method is chosen with a reasonably large step size in order to decrease the computation time. When the tree is close enough to destination the algorithm returns the path and in order to connect the tree to destination point an optimal control problem using single shooting method is defined. To test the algorithm different scenarios are tested and the simulation results show the success of the algorithm.
\end{abstract}

\section{INTRODUCTION}
Path planning has an important role in control of autonomous robots and vehicles. 
In presence of obstacles and constrains on the dynamic and control inputs, it ensures to find a path from initial position to the destination, if any exists, \cite{A1}.
In the last decades, many studies have investigated the motion planning and path planning methods, \cite{A3}, \cite{A4}, \cite{5}, \cite{6}. 
Depending on the problem nature and application, some methods and techniques are more appropriate and work better than others. A vast review on the concept of path planning and existing methods is provided in  \cite{8}, \cite{10}.
Between the existed complete and deterministic methods, when the dimension of the configuration space increases, the computational time grows exponentially. Consequently, complete and deterministic methods usually are not suitable for path planning of high dimensional systems and especially in complex environments which contain complicated obstacles. This has made sampling based motion planning methods more interesting for such problems, \cite{11}. For instance, it has been shown that Rapidly-exploring Random Trees (RRTs) are effective in robotics and autonomous vehicles applications \cite{12}.  
Because of their superior planning capability in high-dimensional space, recently motion planning methods which are based on RRTs have received considerable attention. For example, in \cite{13}, the asymptotically optimal Rapidly-exploring Random Trees -based path planning schemes were proposed. In \cite{14} the authors have used Bezier curves in order to improve the quality of the planned path based on RRTs. Three parallel versions of RRT-based planning methods are compared and a comprehensive survey of sampling- based planning schemes are presented in \cite{15}.
In the recent studies, many researchers have merged and modified the concept of standard RRT to improve its performance, e.g. Goal-biasing-RRT \cite{15}.  Despite these algorithms are not complete, they provide probabilistic completeness in order to ensure that planning is successful as much as possible. Other modified RRT algorithm which are asymptotically optimal are called RRG, RRT*, PRM* algorithms introduced in \cite{16}. It means that, as the number of samples increases and tends to infinity, the path which is obtained by these algorithms converges to the optimal solution with probability one.
LQR-based cost functions are used in \cite{17} for a locally linearized dynamics of non-linear systems, in order to measure distance from destination. \\
In this paper we introduce an approach in order to find a sub optimal path for a non-holonomic system. The method is partially deterministic. In fact this method is similar to rapidly exploring trees, however, the selection of sampling points are wiser instead of choosing random points all around the given environment. To test the algorithm different scenarios are tested and the simulation results show the success of the algorithm.\\
The remainder of the paper is organized as follows. In Section 2, we described the system and the dynamics of vehicles. In addition, it is shown that the 4rth order Runge Kutta has a good stability for even large step sizes and as a trade off between accuracy, computational effort and stability the 4rth order Runge Kutta is a good candidate for our problem.  Then our proposed sampling based method is introduced which is shown that it increases the optimality property compared to standard RRT algorithms. In section 3, simulation results are provided to show the performance and effectiveness of the proposed approach. Section 4 provides the conclusion of this paper. 

\section{Method}
The aim of this study is to introduce an approach in order to find a sub optimal path for a non-holonomic system. The method is partially deterministic. In fact this method is similar to rapidly exploring trees (RRTs), however, the selection of sampling points are wiser instead of choosing random points all around the given environment. The non-holonomic system or vehicle starts to move towards the destination with a cost function. The cost function aims at minimizing the traveled distance and for this purpose a heuristic showing the distance from destination together with traveled distance is used. The bicycle model is used in order to represent the dynamics of the vehicle. After a stability analysis a fourth order Runge Kutta with a step size is used for integration. The selected step size is chosen in a way that the vehicle dynamics is stable and accurate enough to solve this problem. In this section, the vehicle dynamics, search algorithm, stability analysis and optimal control problem are all explained in detail.
\subsection{Vehicle Dynamics}
\begin{center}
\begin{figure}[thpb]
      \centering
   
      \includegraphics[scale=0.4]{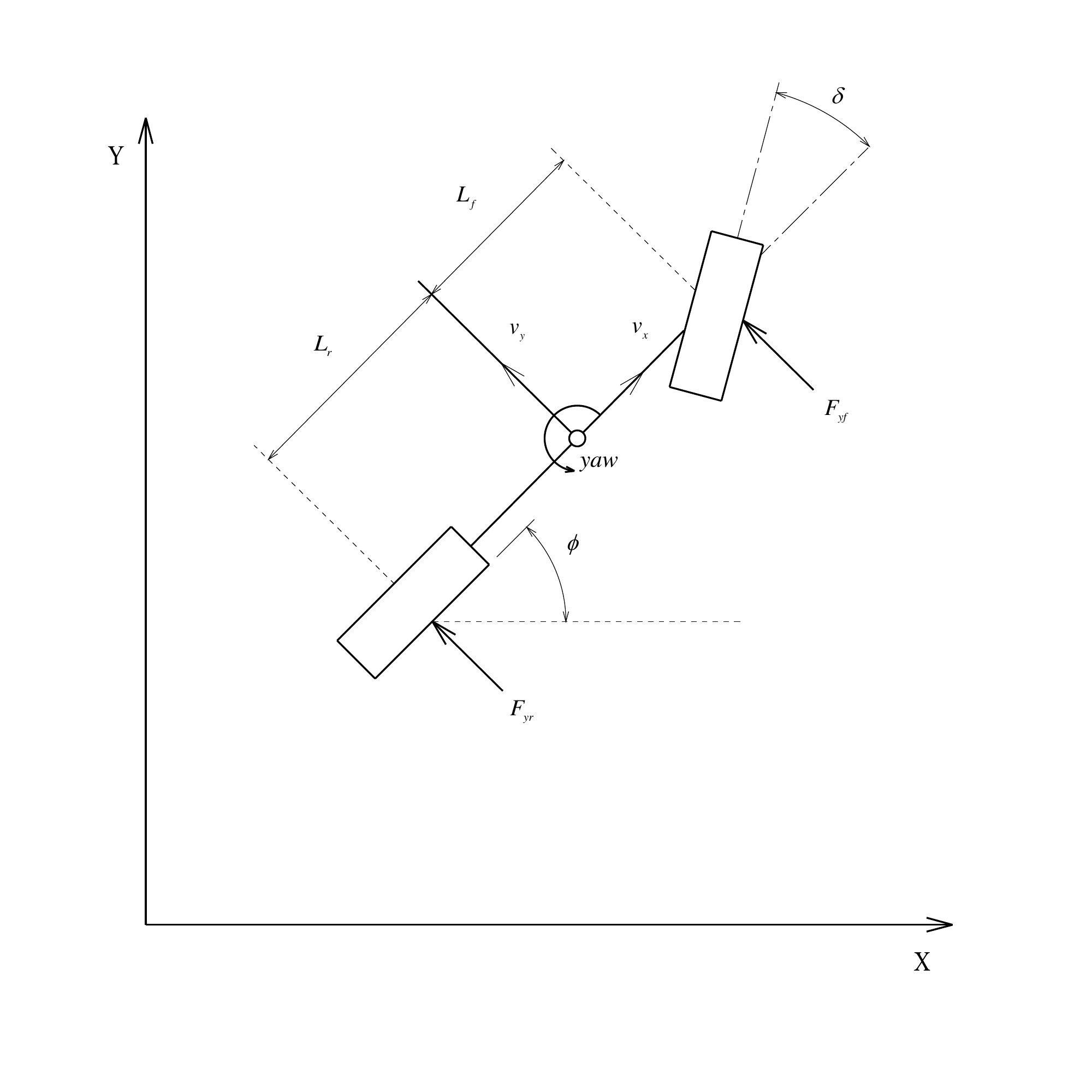}
      \caption{Bycicle dynamics model.}
      \label{bike_dyn}
   \end{figure}
\end{center}

The vehicle dynamics is represented by bicycle model with applied simplifications in order to have less computation time.
The model that is used in this paper is shown in figure \ref{bike_dyn}. The governing dynamics equations for the bycilce model is summerized in equation  \ref{eq_1}, where in this equation $X$ and $Y$ are the global coordinates and and $\theta$ is the heading angle. $vx$ and $vy$ are longitudinal and lateral speeds, respectively and r is the yaw rate. 

\begin{center}   
\begin{flalign}\label{eq_1}
  &&
  \begin{aligned}
\dot X = v_x.cos(\theta) - v_y.sin(theta) \\
\dot Y = v_x.sin(\theta) + v_y.cos(\theta)\\
\dot \theta = r \\
\dot v_x = \frac{-F_{xf}}{m}.cos(\delta) -  \frac{F_{yf}}{m}.sin(\delta) - \frac{F_{xr} }{m}+ v_y.r\\
\dot v_y = \frac{F_{yf}}{m}.cos(\delta) -  \frac{F_{xf}}{m}.sin(\delta) + \frac{F_{yr} }{m} - v_x.r\\
\dot r = \frac{L_f}{I_z}(F_{yf} cos(\delta) - F_{xf} sin(\delta)) -\frac{L_r}{I_z}F_{yr}
  \end{aligned}
  &&
\end{flalign}
\end{center}
In order to include the slip behavior of the vehicle, the slip angles for rear and front wheels are formulated and summerized in in equation \ref{eq_2}. $\delta$ in this equation represents the only input to the vehicle which is steer angle.
\begin{flalign}
  &&
  \begin{aligned}\label{eq_2}
\alpha_f = \frac{v_y+L_fr}{v_x}-\delta\\
\alpha_r = \frac{v_y - L_fr}{v_y}
  \end{aligned}
  &&
\end{flalign}
The slip angles mentioned in equation \ref{eq_2} will result in the lateral forces acting on the vehicle which is calculated by equation \ref{eq_3}.
\begin{flalign}
  &&
  \begin{aligned}\label{eq_3}
F_{yf} = -C_{\alpha f}.\alpha_f\\
F_{yr} = -C_{\alpha r}.\alpha_r
  \end{aligned}
  &&
\end{flalign}
As mentioned previously, in order to reduce computation time the states of the vehicle model is reduced. For this purpose some simplifications were made like assuming a constant longitudinal speed and no aerodynamic forces. After simplifications applied to equation \ref{eq_1} the vehicle dynamics is reformulated as equation \ref{eq_4}.
\begin{flalign}
  &&
  \begin{aligned} \label{eq_4}
\dot X = v_x.cos(\theta) - v_y.sin(theta) \\
\dot Y = v_x.sin(\theta) + v_y.cos(\theta)\\
\dot \theta = r \\
\dot v_y = \frac{F_{yf}}{m}.cos(\delta) -  \frac{F_{xf}}{m}.sin(\delta) +\frac {F_{yr}}{m} - v_x.r\\
\dot r = \frac{L_f}{I_z}(F_{yf} cos(\delta)  - \frac{L_r}{I_z}F_{yr}
  \end{aligned}
  &&
\end{flalign}
\subsection{Stability of integration method}
\begin{figure}[H]
   
      \includegraphics[scale=0.9]{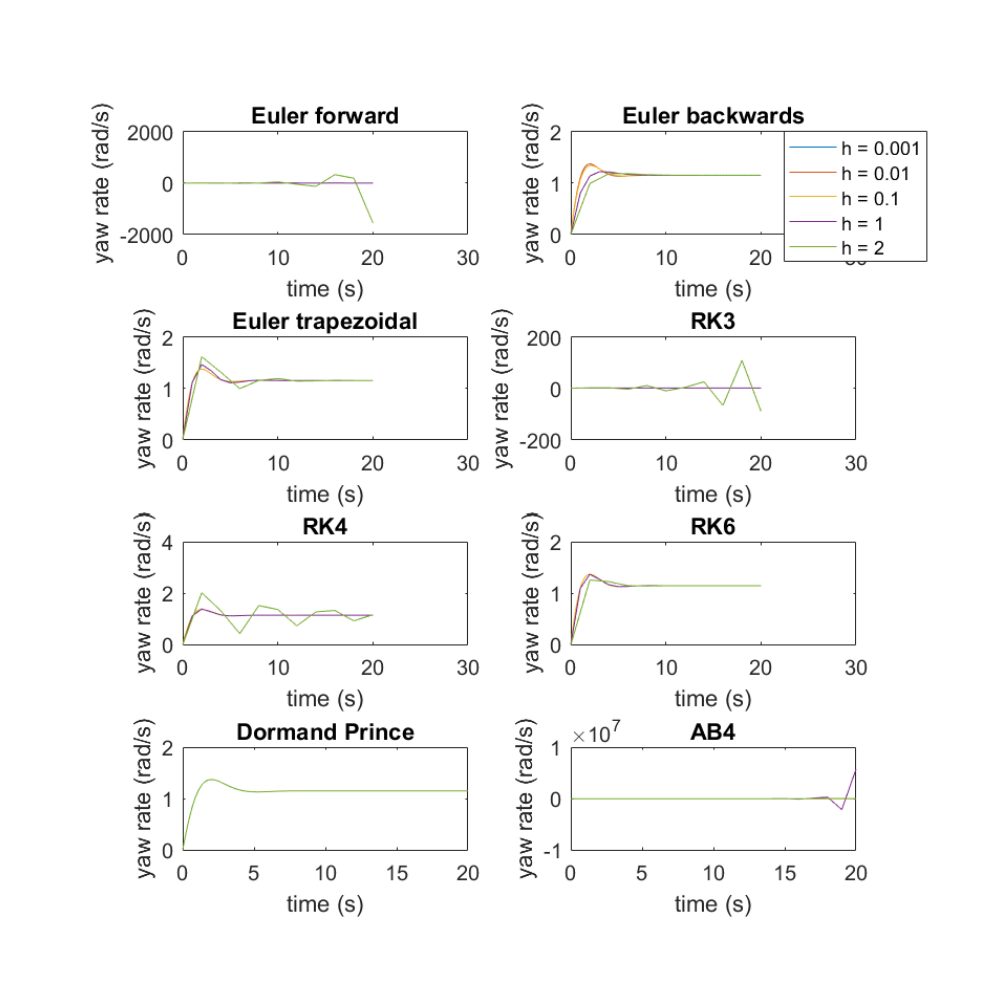}
      \caption{Stability test of integration methods in calculation of yaw rate.}
      \label{RK_YAW}
   \end{figure}

\begin{center}
   \begin{figure}[H]
   
      \includegraphics[scale=0.9]{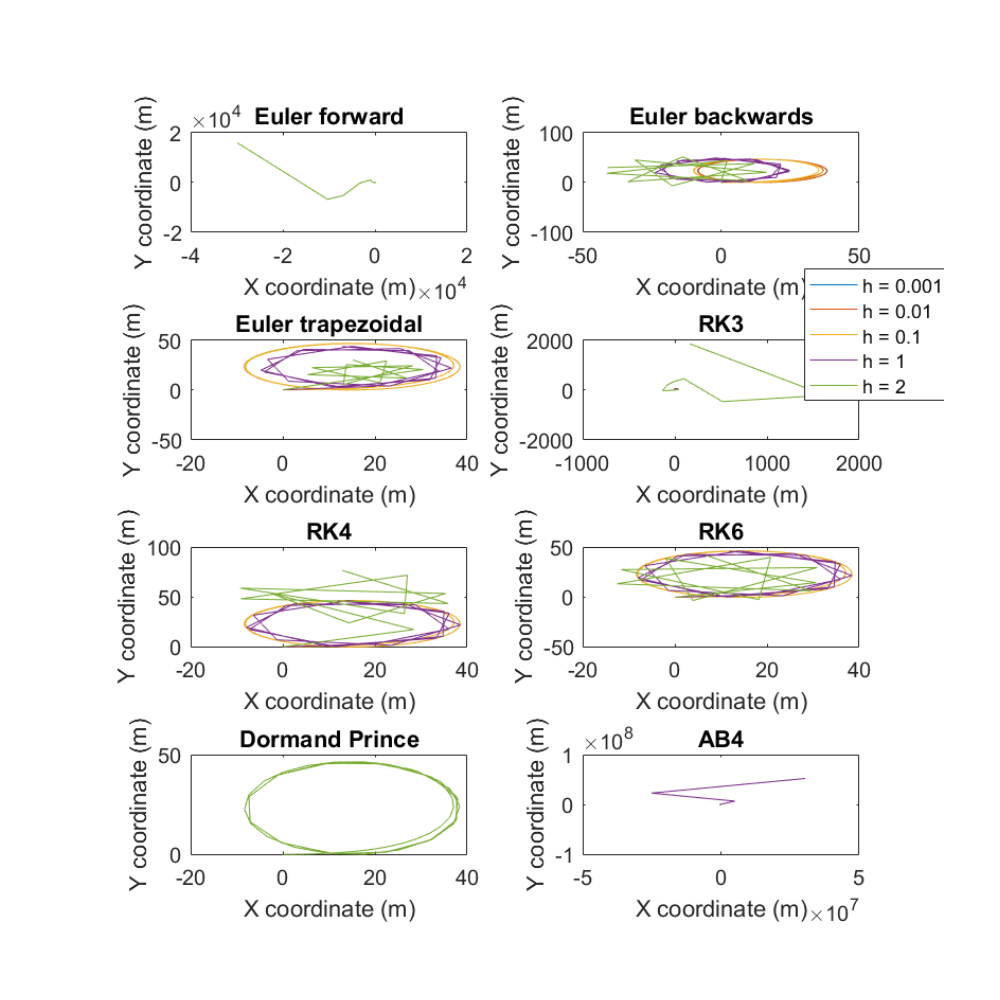}
      \caption{Stability test of integration methods in calculation of global coordinates.}
      \label{RK_XY}
   \end{figure}
\end{center}

The vehicle dynamics represented by bicycle model, should be integrated at each iteration during the search algorithm. For this purpose the integrator should be accurate, stable and fast enough in order to reduce calculation time. For this purpose, the bicycle dynamics discussed in previous section is integrated with different methods in finite time. The integration methods that are used during the test are, Euler Forward, Euler Backwards, Euler Trapezoidal, 3rd order Runge-Kutta, 4th order Runge-Kutta, 6th order Runge-Kutta, Dormand-Prince method and 4th order Adams-Bashforth methods. \\
The test was performed with a constant longitudinal speed of $15 m/s$ and a steer angle of $pi/4$. 
The Dormand-Prince method is an adaptive method that is used to illustrate real values of the parameters and make comparison with other methods. In order to show the stability of different methods, only the states with higher oscillations and unstability were selected to illustrate. \\
As shown in the figure \ref{RK_YAW} and \ref{RK_XY}, Euler Forward, 3rd order Runge-Kutta and Adams-Bashforth methods are not stable methods for large step sizes. Among the other methods, the 4rth order Runge Kutta shows a good stability for even large step sizes and as a trade off between accuracy, computational effort and stability the 4rth order Runge Kutta is a good candidate for this problem. 
%

%

%
%


\subsection{Sampling Based Search Method}
In this section the sampling based method used for search algorithm is explained. The main pseudo code that is used to find the path is shown in algorithm \ref{main}.
\begin{center}
   \begin{figure}[H]
      \centering
   
      \includegraphics[scale=0.6]{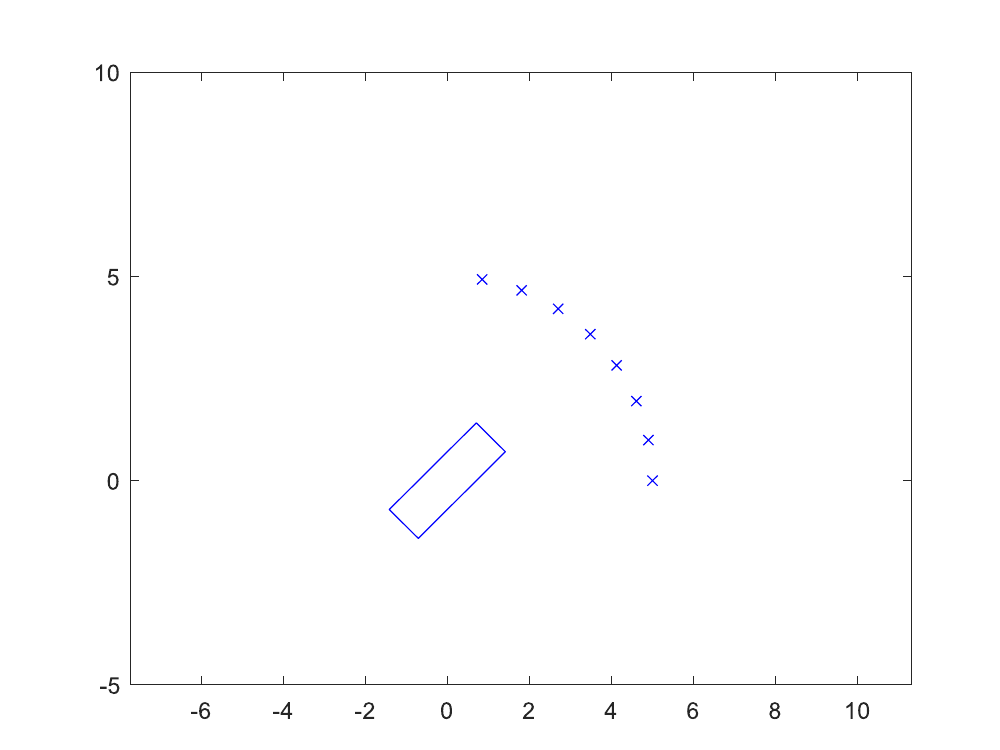}
      \caption{The position of sampling points with respect to vehicle position.}
      \label{sample}
   \end{figure}
\end{center}   

\begin{center}
\begin{algorithm}
\caption{Semi optimal path planning}\label{euclid}
\label{main}
\begin{algorithmic}[1]
\Procedure{PathFinder}{}
\State $\textit{Initialize-tree (start, dest, map)} $
\State $\textit{Initialize-StateMatrix} $
\State $\textit{Initialize-parent vector} $
\While {dist $>$ mindist}{
\State $current \gets \textit{Tree point with min cost}$
\State $currentState \gets \textit{StateMatrix[current]}$
\State $SamplingPoints \gets \textit{PointSelector(currentState)}$
\State $Biased rand points \gets \textit{rand points around dest}$
\State $AllPoints \gets \textit{Sampling points + Biased rand points}$
\State $TreePointsCosts\gets \textit{Shooting(currentState, AllPoints)}$
\For {all shooted points}
\If {$i > \textit{cost ~= inf}$} {
Add points to tree
}
\EndIf
\EndFor

}
\EndWhile
\State $dist \gets \textit{minDist(tree,dest)}$.
\EndProcedure
\end{algorithmic}
\end{algorithm}
\end{center}
 In this method instead of randomly selecting one point in the configuration space, several points are selected but not randomly. Since the location and states of the start point is known in advance, it is obvious that only the front of the vehicle is useful to sample (figure \ref{sample}). Therefore, it is reasonable to sample a region between $\theta - \delta_{max}$ and $\theta + \delta_{max}$ ($\delta_{max}$ is maximum steer angle). Because, even if a point is sampled out of this region the vehicle will take maximum steer angle either to the left or to the right. Algorithm \ref{selection} shows how the selection of sampling points takes place. After sampling the front of the vehicle the proper steer angle to move in that direction is calculated and fed into the vehicle dynamics.

\begin{center}
\begin{algorithm}
\caption{Selection of Points}\label{euclid}
\label{selection}
\begin{algorithmic}[1]
\Procedure{PointSelector}{}
\State $\textit{X} \gets \text{currentState[X] }$
\State $\textit{Y} \gets \text{currentState[Y] }$
\State $\textit{HeadAngle} \gets \text{currentState[HeadAngle] }$
\State $\textit{Arc} \gets \text{Generate a arc with respect to X, Y, HeadAngle}$
\State $\textit{Points} \gets \text{Choose n equally spaced points on arc}$\\
\Return{Points}
\EndProcedure

\end{algorithmic}
\end{algorithm}
\end{center}
As shown in algorithm \ref{shooting}, after finding the steer angle the integration takes place for a given amount of time horizon and for all the integrated points a cost is calculated. The cost function includes a heuristic which is distance from the destination and traveled distance until integrated point. The two calculated costs will add up together and give a single cost value for all integrated points. If the integrated points have collision with obstacles the cost is set to infinite. \\
The cost value will be fed to the main algorithm  (algorithm \ref{main}). The next start point for the algorithm will be a point with minimum cost. The algorithm keeps calculating until it reaches to the goal region.
\begin{center}
\begin{algorithm}
\caption{Shooting of states}\label{euclid}
\label{shooting}
\begin{algorithmic}[1]
\Procedure{Shootings}{}
\For{all sampling points}
\State $\textit{Calculate SteerAngle}$
\State $\textit{new Points} \gets \text{Integration(CurrentState,SteerAngle)}$
\State $\textit{H} \gets \text{calculate heuristic for all new points}$
\State $\textit{g} \gets \text{calculate traveled distance for all new points}$
\If {$collison$} {H=inf}
\EndIf
\EndFor\\
f=g+H\\
\Return{f}
\EndProcedure
\end{algorithmic}
\end{algorithm}
\end{center}
\subsection{Optimization Problem}
As mentioned in previous section when the tree reaches the goal region the algorithm stops searching. Because of non-holonomic system it is not possible to directly connect the nearest point of the tree to the goal point with a line. Instead an optimization method is used to connect the nearest point to the goal point. Using single shooting method all the states are shooted with an integration method and a vehicle dynamic. For simplicity the vehicle dynamics is a point mass model. The point mass dynamics is represented in equation \ref{pm}. The states and inputs of the vehicle dynamics are bounded and summarized in the optimization problem in equation \ref{opt}. The optimization software minimizes the total time required to go from nearest point of the tree to the goal point.
\begin{center}   
\begin{flalign}
  &&
  \begin{aligned}\label{pm}
\dot X = v.cos(\theta) \\
\dot Y = v.sin(\theta) \\
\dot \theta = v.tan(\delta)/L \\
\dot v = \frac{F}{m}\\
  \end{aligned}
  &&
\end{flalign}
\end{center}
\begin{equation}\label{opt}
\begin{matrix}
\displaystyle \min & t_f & \\
\textrm{s.t.} & X_{min} &  \leq & X & \leq & X_{max} \\
& Y_{min} &  \leq & Y & \leq & Y_{max}  \\
&-\pi &  \leq & \theta & \leq & \pi\\
&0 &  \leq & v & \leq & 25\\
&-inf &  \leq & F & \leq & inf\\
&-\pi/4 &  \leq & \delta & \leq & pi/4\\
&x_{n+1} &  = & x_n + f(x).dt\\
&dt &  = &t_f/N \\
& [X_0, Y_0, \theta_0, v_0] & = & S_{initial}\\
& [X_f, Y_f, \theta_f, v_f] & = & S_{final}\\
\end{matrix}
\end{equation}
In order to let the vehicle find the path with respect to vehicle dynamics the states are shooted in different trajectories using single shooting method. The shooted states should meet the constraints mentioned in the optimal control problem.\\
Thereafter the path computed by optimal control problem is added to the main tree and this forms the whole path from start point to the destination point.
\section{Results and Discussin}
\begin{center}
   \begin{figure}[H]
      \centering
   
      \includegraphics[scale=0.6]{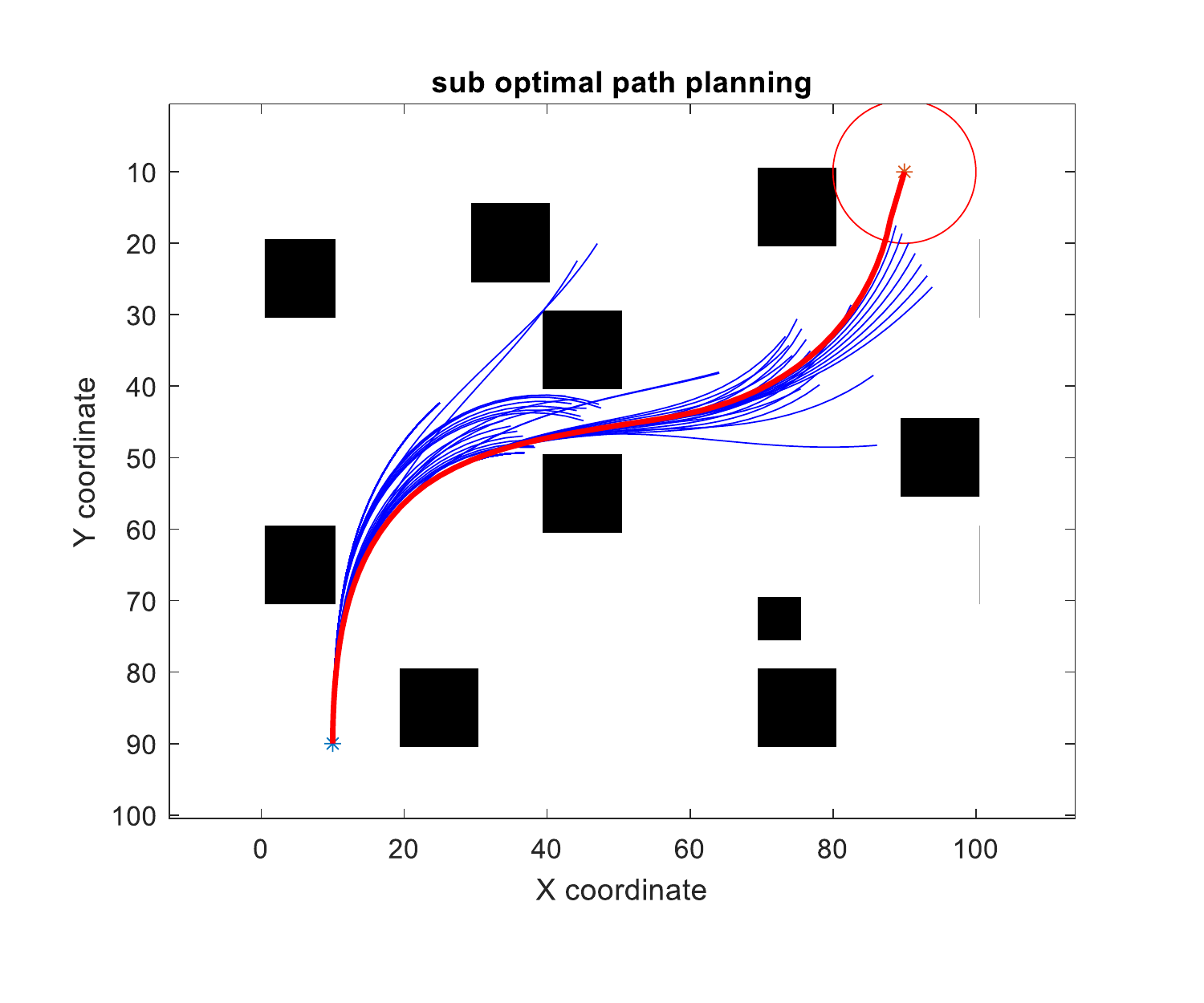}
      \caption{Map1}
      \label{Map2}
   \end{figure}
\end{center}

\begin{center}
\begin{figure}[H]
      \centering
   
      \includegraphics[scale=0.6]{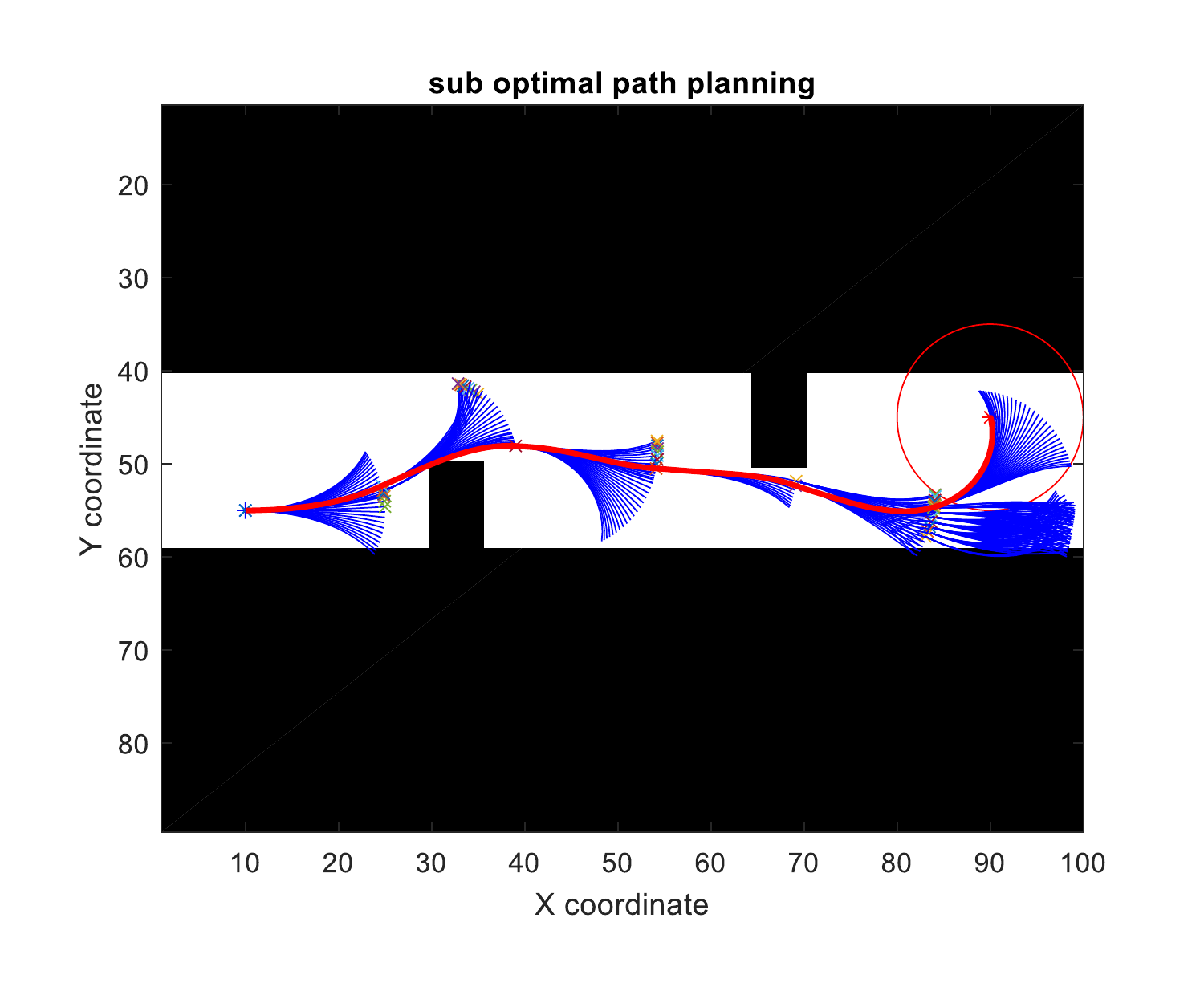}
      \caption{Map4}
      \label{Map3}
   \end{figure}
\end{center}

The previously mentioned algorithms are tested in different scenarios with different obstacles where some of them are discussed in this section. In the figures discussed in this section the black regions are obstacles and white regions are the free spaces that vehicle can travel. The blue curves represent the explored tree during search method and the red curve shows the final path computed by the algorithm. Figure \ref{Map2} illustrates  a configuration space with some obstacles between start point and destination point. 
Figure \ref{Map3} shows another case where the configuration space is similar to real street but with added obstacles. During the simulations it was noted that the algorithm returns almost same path after repeating it many times since the method is not probabilistic. Another advantage with this method is that it finds the path in a small amount of time.
\\

\begin{center}
\begin{figure}[H]
      \centering
   
      \includegraphics[scale=0.6]{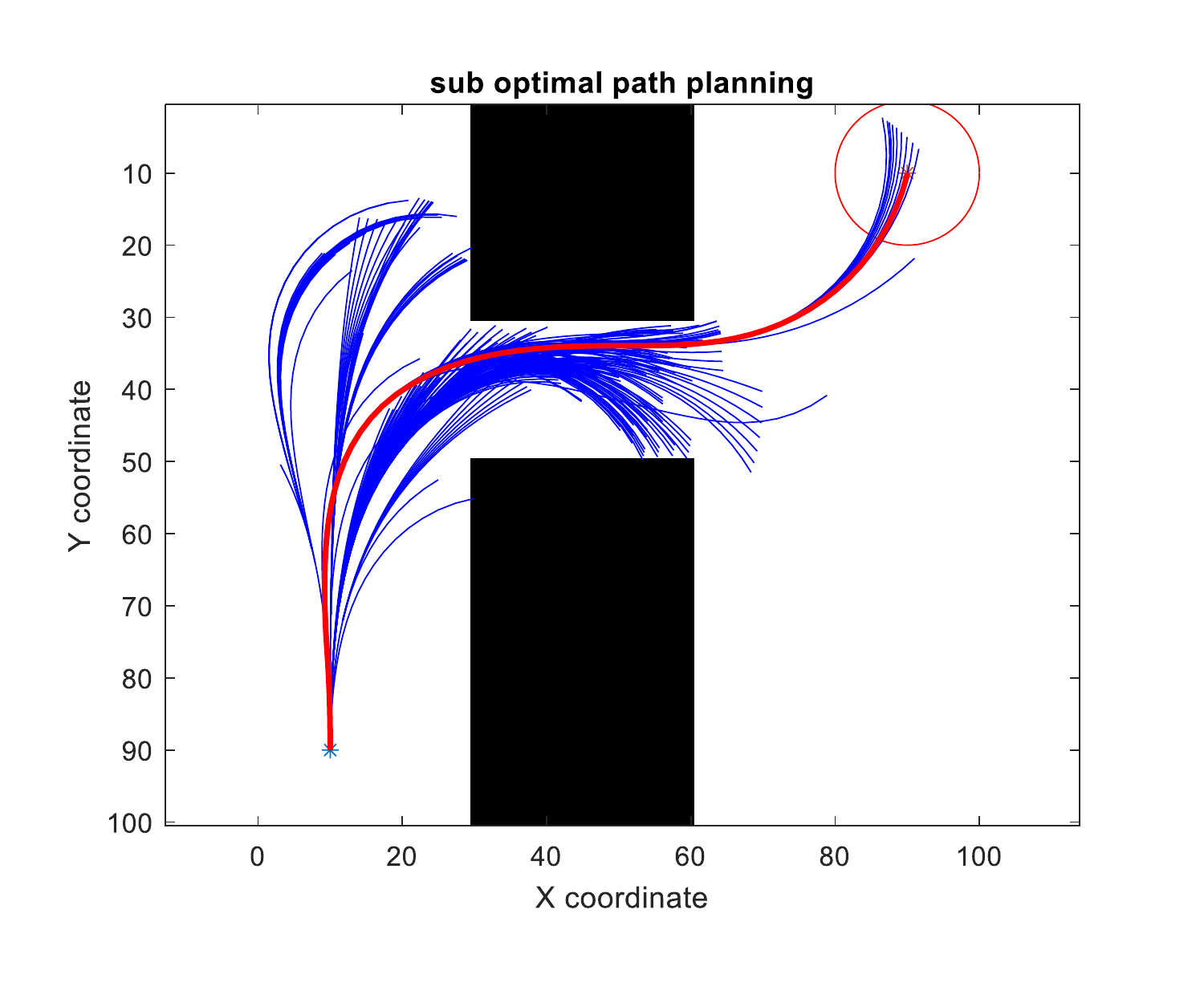}
      \caption{Map2}
      \label{Map1}
   \end{figure}
 \end{center}

\begin{figure}[H]
      \centering
   
      \includegraphics[scale=0.6]{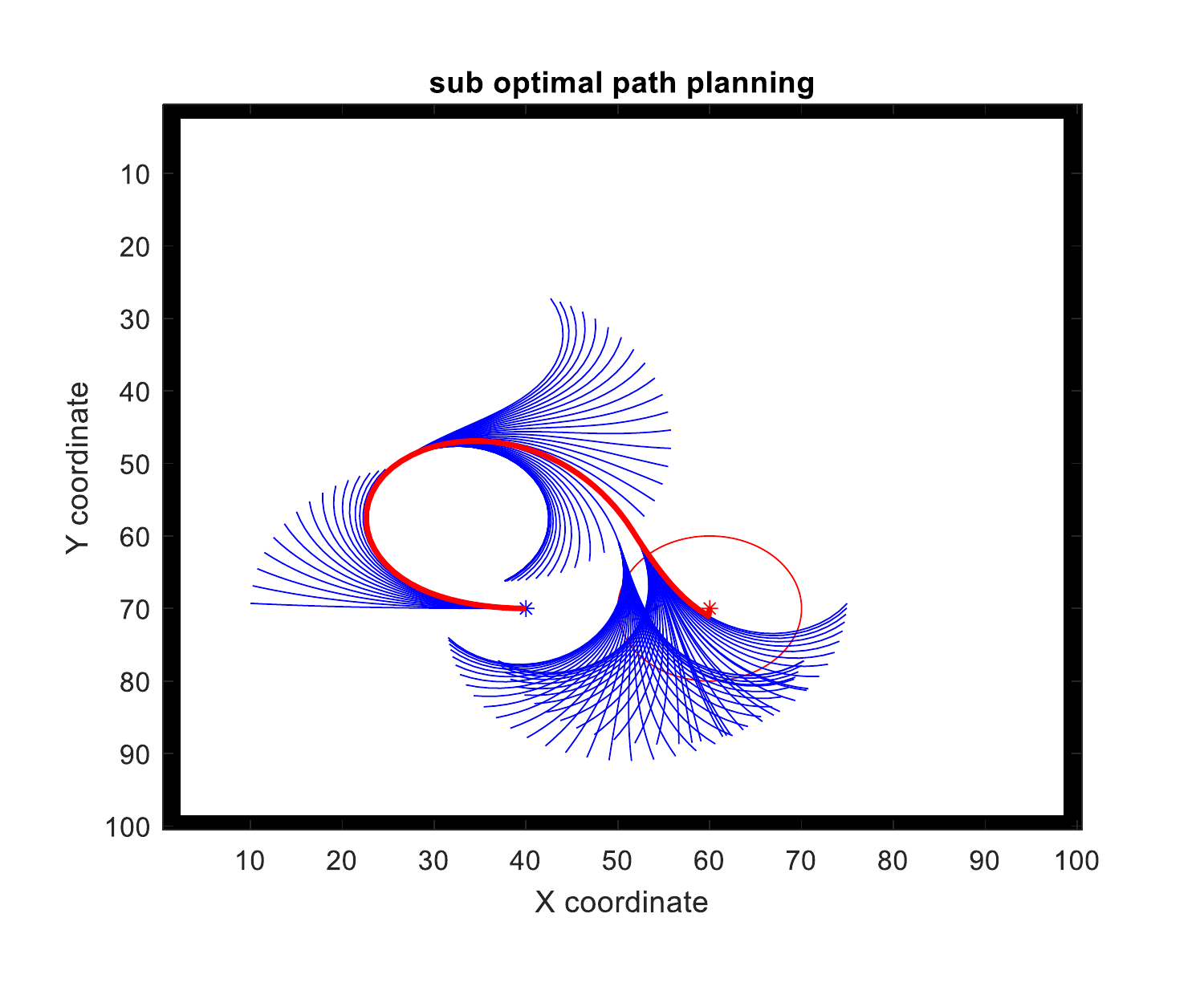}
      \caption{Map3}
      \label{Map5}
   \end{figure}

\begin{figure}[H]
      \centering
   
      \includegraphics[scale=0.6]{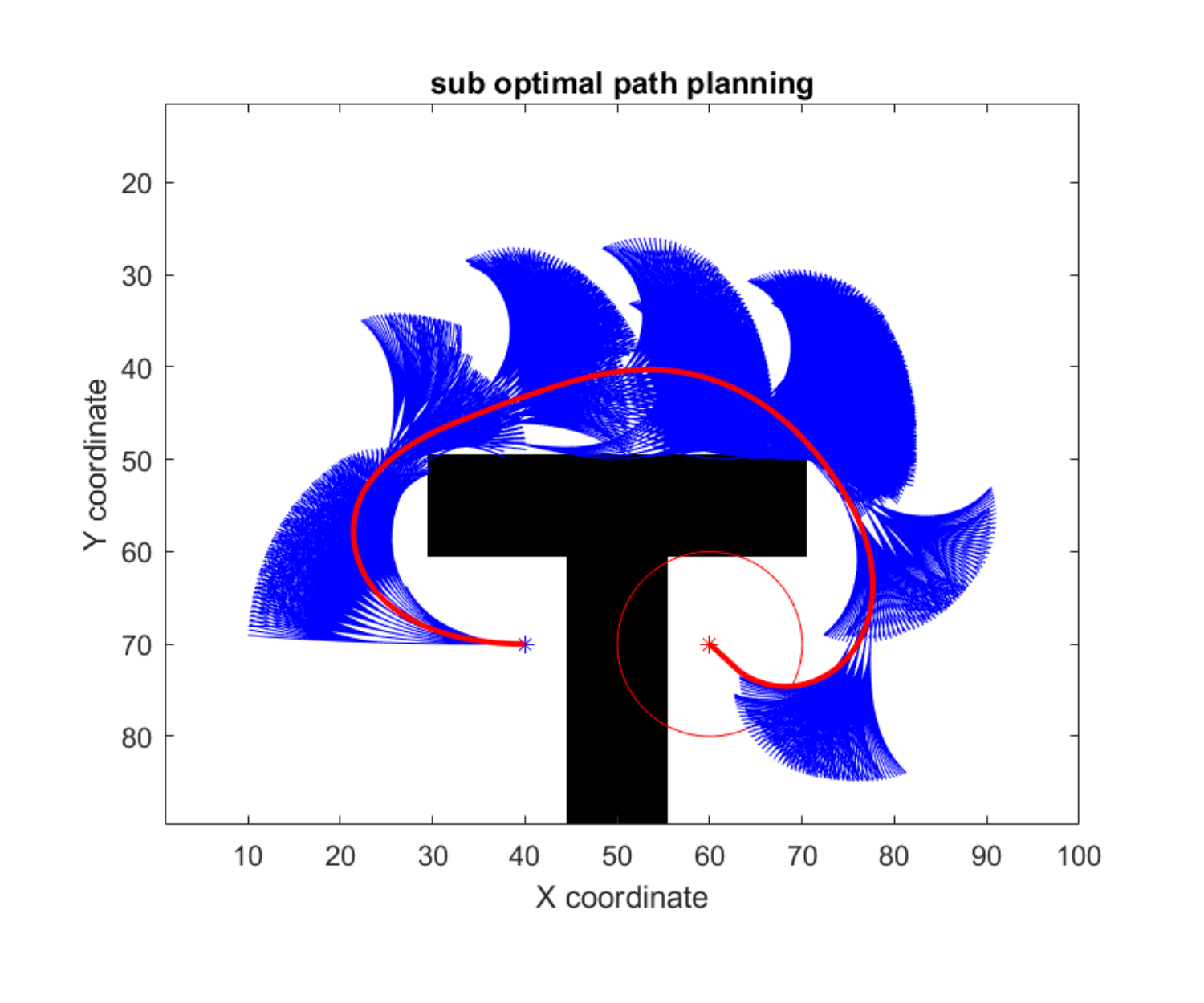}
      \caption{Map5}
      \label{Map4}
   \end{figure}

The figures \ref{Map1} and \ref{Map5} show other cases with larger free space for the algorithm to explore. As depicted in this figure the algorithm explores only a small region in configuration space.
 Another simulation result with different configuration space is also depicted in the figure \ref{Map4} where there is an obstacle between the start point and destination point. The algorithm manages to find the solution for this case also but since the algorithm aims at minimizing the traveled distance it takes a while to explore the suitable path.
 
\section{CONCLUSIONS}
The method proposed in this paper aims at finding a sub-optimal path for the vehicle. The path can be further optimized by using many sampling points. However, there is a trade off between computation time and a better path in terms of optimality The method successfully returns a path within small computation time and it turns out to be the same result after repeating the algorithm. The computation time is further reduced by doing a stability analysis on the integration methods and choosing a reasonably large step size.
To test the algorithm different scenarios were tested and the simulation results have showed the success of the algorithm. 


\bibliographystyle{IEEEtran}    
 \bibliography{IEEEtran}

\end{document}